\documentclass[runningheads]{style/llncs}
\usepackage{graphicx}
\usepackage{multirow}
\usepackage{booktabs}
\usepackage{subcaption}
\usepackage{wrapfig}
\usepackage{sidecap}
\usepackage{colortbl}

\usepackage{style/comment}
\usepackage{amsmath,amssymb} 
\usepackage{color}

\usepackage[accsupp]{axessibility}  

\begin{document}
\pagestyle{headings}
\mainmatter
\def\ECCVSubNumber{1316}  

\title{Unsupervised Visual Representation Learning by Synchronous Momentum Grouping} 


\titlerunning{Unsupervised Visual Representation Learning by SMoG}
%
\author{Bo Pang\inst{1}\and
Yifan Zhang\inst{1} \and
Yaoyi Li\inst{2} \and
Jia Cai \inst{2} \and
Cewu Lu\inst{1}\thanks{Cewu Lu is the corresponding author.}}
\authorrunning{B. Pang et al.}
%
\institute{Shanghai Jiao Tong University\\
\email{\{pangbo, zhangyf\_sjtu, lucewu\}@sjtu.edu.cn}\and
HuaWei Technologies Co., Ltd.\\
\email{\{liyaoyi, caijiai1\}@huawei.com}\\}

\maketitle

\begin{abstract}
In this paper, we propose a genuine group-level contrastive visual representation learning method whose linear evaluation performance on ImageNet surpasses the vanilla supervised learning. Two mainstream unsupervised learning schemes are the instance-level contrastive framework and clustering-based schemes. The former adopts the extremely fine-grained instance-level discrimination whose supervisory signal is not efficient due to the false negatives. Though the latter solves this, they commonly come with some restrictions affecting the performance. To integrate their advantages, we design the SMoG method. SMoG follows the framework of contrastive learning but replaces the contrastive unit from instance to group, mimicking clustering-based methods. To achieve this, we propose the momentum grouping scheme which synchronously conducts feature grouping with representation learning. In this way, SMoG solves the problem of supervisory signal hysteresis which the clustering-based method usually faces, and reduces the false negatives of instance contrastive methods. We conduct exhaustive experiments to show that SMoG works well on both CNN and Transformer backbones. Results prove that SMoG has surpassed the current SOTA unsupervised representation learning methods. Moreover, its linear evaluation results surpass the performances obtained by vanilla supervised learning and the representation can be well transferred to downstream tasks.
\end{abstract}

\section{Introduction}
In the era of adopting deep learning and data-driving as the mainstream framework~\cite{lecun2015deep}, the quality of the learned representation largely determines the performance of models on a majority of tasks~\cite{zhou2016learning}. For a long time, people utilize supervised tasks and adopt a large amount of annotations to train models and get good representations. Nevertheless, this simple and efficient scheme faces the problems of expensive and time-consuming annotating costs~\cite{asano2020labelling,fabbri2018learning,fang2019instaboost,sun2022correlation}, and non-ideal generalization performance caused by bias from the annotation information~\cite{ben2007analysis,pan2010domain}. To solve these problems, people extensively study the unsupervised representation learning method~\cite{sun2022human}, including the generative models providing expressive latent features~\cite{kingma2013auto,rezende2014stochastic} and self-supervised methods with different pretext tasks~\cite{vincent2008extracting,vincent2008extracting,pathak2016context,noroozi2016unsupervised}. However, there are still performance gaps between these methods and the supervised scheme. In recent years, unsupervised representation learning achieves great improvements, SOTA contrastive learning~\cite{simclr,moco,swav,byol} based unsupervised methods reduce the performance gap to less than 3\%. We continue the study of the unsupervised representation learning, stand on the shoulder of previous contrastive and clustering-based methods, and finally, push the performance of unsupervised methods above the level of the vanilla supervised scheme.

\begin{wrapfigure}[28]{r}{0.5\linewidth}
	\vspace{-0.32in}
	\includegraphics[width=\linewidth]{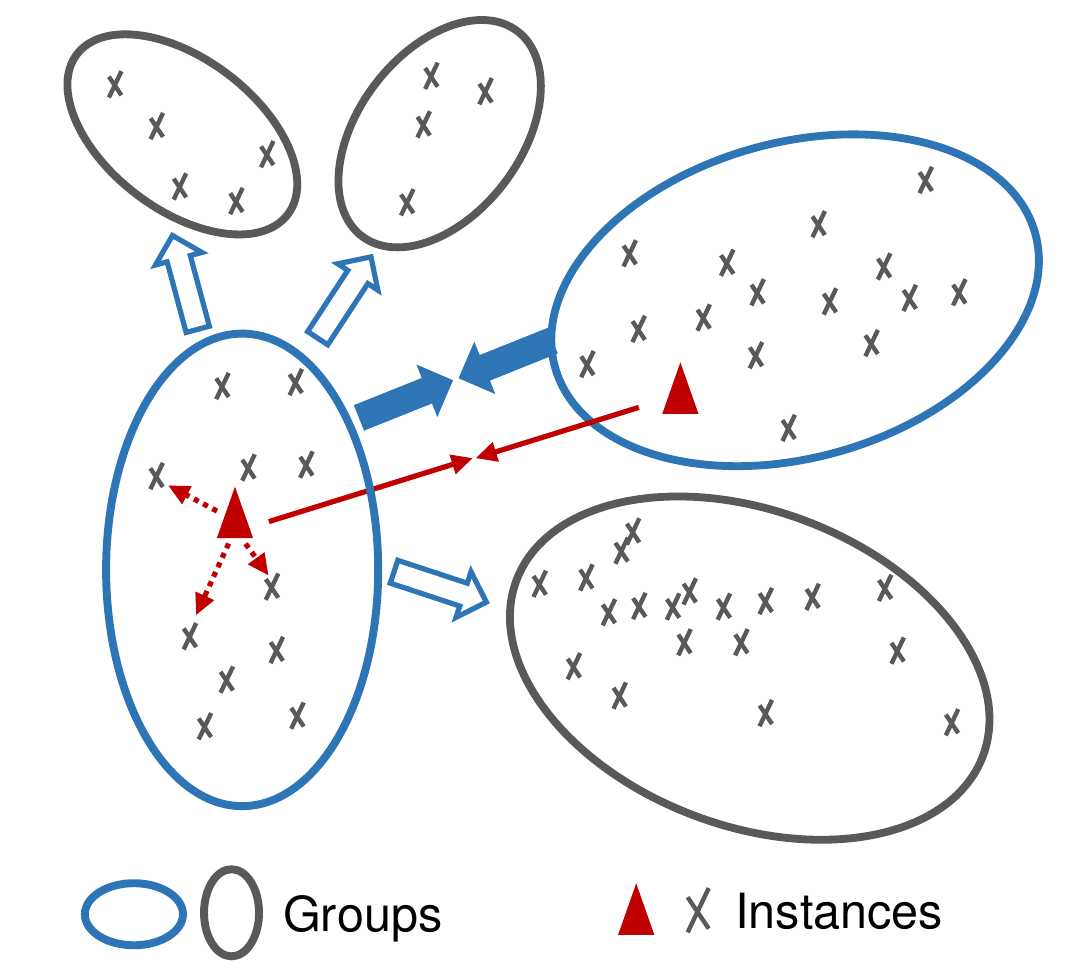}
	\caption{Group contrastive \textit{vs.} instance contrastive learning. Compared with instance contrasting (red parts), group contrasting (blue part) learns representations through higher-level semantics, which can reduce the chance that the already learned similar instances are still treated as negative pairs (false negatives). The significant reduction of contrast elements after the grouping process also makes global contrasting easier to calculate. The colorized instances or groups denote the positive pairs and the grey ones are negatives.}
	\label{fig:cover}
\end{wrapfigure}

Traditional contrastive learning methods adopt instance discrimination as the pretext task. This kind of refinement of classification treats every sample as a category to conduct discrimination. Thus, it will introduce many false-negative pairs, leading to inefficient supervisory signals. And another problem is that the accurate instance discrimination requires pair-wise comparison on the entire dataset. However, with limited resources, most implementations compromise to adopt a subset of the comparison with a large batch size or memory bank~\cite{moco,wu2018unsupervised}, which further decreases the efficiency of learning.

Previous clustering-based methods relax instance discrimination into group discrimination to solve these problems. The core task of clustering is how to split instances into groups. They try to adopt K-means clustering method~\cite{deepcluster,ODC} or optimal transport~\cite{sela,swav} method to achieve this, but due to the limitations they introduce (asynchronous two-stage training or local equipartition respectively), they don't show an advantage in terms of performance.

Here, we integrate the advantages of instance contrastive and clustering-based methods to the newly proposed SMoG method which follows the instance contrastive framework, and extend it to group-level (see Fig.~\ref{fig:cover}). One major design is the momentum grouping scheme that allows gradient propagating through groups to synchronously conduct the instance grouping process together with the representation learning. It is the first method to apply contrasting directly on groups instead of instances, since no gradient can propagate through group features in previous methods, With it, SMOG avoids the false negatives in instance contrasting and wipes off limiting factors in previous clustering-based methods.

The proposed SMoG method is effective and pretty simple.  We evaluate it on several standard self-supervised benchmarks and multiple downstream tasks. In particular, under the linear protocol, it achieves 76.4\% top-1 accuracy on ImageNet with a standard ResNet-50~\cite{resnet}, which surpasses the vanilla supervised-level performance for the first time! Extensive experiments show that SMoG works robustly and we observe consistent SOTA performance under linear (+0.8\%), semi-supervised classification (+2\%), detection, and segmentation tasks. Besides, SMoG works well on both CNN and Transformer backbones.

\section{Related Work}
\subsection{Handcraft Pretext Task} 
Adopting pretext tasks to provide training supervisory signals is an effective unsupervised representation learning scheme. By now, well studied pretext tasks include jigsaw solving~\cite{noroozi2016unsupervised,noroozi2018boosting}, colorization~\cite{vincent2008extracting,larsson2017colorization,zhang2016colorful}, denoising~\cite{vincent2008extracting}, inpainting~\cite{pathak2016context}, super-resolution~\cite{ledig2017photo}, and patch position prediction~\cite{doersch2015unsupervised,doersch2017multi}. In addition, for video visual inputs with temporal dimension, sequential-related pretext tasks are proved useful, such as ordering frames~\cite{fernando2017self,misra2016shuffle,wei2018learning}, motion estimation~\cite{agrawal2015learning,jayaraman2015learning,DBLP:journals/corr/abs-1712-01337,liu2018switchable}, and further frame estimation~\cite{lotter2016deep,mathieu2015deep,srivastava2015unsupervised,vondrick2016anticipating,vondrick2016generating}. Due to the specificity of these tasks, the learned representation often carry a certain bias, leading to relatively limited performances on downstream tasks.

\subsection{Instance Discrimination Method}
Present mainstream contrastive learning methods~\cite{moco,simclr,byol,swav,ressl,HSIC,Relic,care,unigrad,pang2021unsupervised} learn representations by instance discrimination which treats each image (pixel) in a training set as its own category. This scheme achieves state-of-the-art performances on downstream tasks. Its common training direction provided by InfoNCE~\cite{cpc,simclr,gutmann2010noise} or its variations is to maximize the mutual information~\cite{cpc,informax}, which requires a large number of negative pairs to get good performances~\cite{simclr}. Adopting a large batch size is a straightforward method, but it consumes lots of resources. To solve this, MoCo~\cite{moco} and \cite{wu2018unsupervised} propose to utilize memory structures. Some latest designs~\cite{byol,simsiam,dino} conduct contrasting without negative samples by an asymmetry Siamese structure or normalization techniques. Recent work~\cite{HSIC} proposes self-supervised learning with the Hilbert-Schmidt Independence Criterion, which yields a new understanding of InfoNCE. While RELIC~\cite{Relic} improves the generalization performance through an invariance regularizer under the causal framework. NNCLR~\cite{nnclr} adopts the nearest neighbour from the dataset as the positives. UniGrad~\cite{unigrad} provides a unified contrastive formula through gradient analysis. Instance discrimination, as an excessive fine-grained classification setting implies some problems like false-negative pairs~\cite{zhuang2019local}.

\subsection{Group Discrimination Method}
Our work follows the group discrimination scheme. DeepClustering~\cite{deepcluster} adopts the K-means clustering method to get the groups and learn features from them in an iterative manner. However, the two-step training scheme (clustering, learning) leads to delayed supervisory signal, against effective representation learning. ODC~\cite{ODC} and CoKe~\cite{coke} shortens the two-step circulation period but does not solve the delay problem. SeLa~\cite{sela} and SwAV~\cite{swav} treat the grouping problem as pseudo-labelling and solve it as a optimal transport task. But, to avoid degeneracy, they add an equipartition constraint, decreasing the validity of grouping. Our work stands on the shoulders of these works, solves the mentioned problems, and finally achieves the level of vanilla supervised learning for the first time.

\section{Approach}
Present SOTA methods commonly adopt the instance contrastive learning, apply a two-stream Siamese network structure, and take InfoNCE or its variants as the loss function to conduct the contrasting. This is an eminent and robust structure that has been proved by many models. However, there are two potential problems: 1) In terms of the problem setting, instance discrimination is too fine-grained and the learned representation goes against the downstream tasks that rely on the high-level semantics to some extent, because for high-level semantics, instance-level contrasting introduces many false-negative pairs which damage the quality of representation learning. 2) In terms of technical practice, contrastive learning based on InfoNCE needs a large number of negative pairs to improve the upper bound of the theoretical performance~\cite{simclr,moco}, which poses a challenge on computing resources or needs specific model designs. 

To deal with the above two problems, previous works~\cite{swav,deepcluster,cld} propose the clustering-based method: adopting groups as negatives to reduce false-negatives and the total number of negative pairs. However, these methods introduce some limitations and lose the advantages of contrastive methods: the always-updating optimization signal and totally unrestricted feature distribution. To integrate the advantages of contrastive and clustering-based method, we propose to contrast in the group unit and design the first group-level contrastive learning algorithm Synchronous Momentum Grouping (SMoG) which inherits and develops current contrastive learning techniques, and pushes the performance of unsupervised methods to the level of vanilla supervised scheme.

\subsection{Group-Level Contrastive Learning}
In general, given a dataset $\mathbf{X}=\{\mathbf{x}_1, \mathbf{x}_2, ...\mathbf{x}_n\}$ with $n$ instances and a visual model $f_\theta$ with parameter set $\theta$ which maps each instance to a vector representation $f_\theta(\mathbf{x})$, the pipeline of the instance contrastive learning framework can be expressed as:
\begin{equation}
L_i = -{\rm log}\frac{{\rm exp}({\rm sim}(f_\theta(\mathbf{x}_i^a), \hat f_\eta(\mathbf{x}_i^b) /\tau)}{\sum_{\mathbf{x}_j\neq \mathbf{x}_i^a, \mathbf{x}_j\in \mathbf{Y}\subseteq \mathbf{X}}{{\rm exp}({\rm sim}(f_\theta(\mathbf{x}_i^a), \hat f_\eta(\mathbf{x}_j) )/\tau)}},
\end{equation}
where ${\rm sim}(\mathbf{u}, \mathbf{v})$ is commonly instantiated as the normalized inner product, $\hat f_\eta$ is just $f_\theta$ or a variant of $f_\theta$ such as its momentum updated version or the version without the last few layers (predictor). $\mathbf{x}_i^a$ and $\mathbf{x}_i^b$ are two different augmented views of $\mathbf{x}_i$. In theory, $\mathbf{x}_j$ should come from the training set $\mathbf{X}$, but in practical, taking computing complexity into account, $\mathbf{x}_j$ is commonly selected from a much smaller subset $\mathbf{Y}\subseteq \mathbf{X}$.

In order to extend this framework to the group level, we need to first define and generate a certain number of groups with group feature $\mathbf{g}$, attach each instance to a certain group, allow gradients propagating through groups, and update them synchronously during training: $\mathbf{c}_i^a, \{\mathbf{g}\} \leftarrow \phi(f_\theta(\mathbf{x}_i^a) | \{\mathbf{g}\})$, where $\phi$ is the group assigning function that takes in instance and group features, generates instances' corresponding group $\mathbf{c}_i^a$, and updates the groups $\{\mathbf{g}\}$. Thus, $\mathbf{c}_i = \mathbf{g}_k$ means that instance $\mathbf{x}_i$ is attached to group $k$. With the group-level features, we can derive the group-level contrastive learning framework as:
\begin{equation}
L_i = -{\rm log}\frac{{\rm exp}({\rm sim}(\mathbf{c}_i^a, \mathbf{c}_i^b)/\tau)}{\sum_{\mathbf{g}_j \in \mathbf{G}}{\rm exp}({\rm sim}(\mathbf{c}_i^a, \mathbf{g}_j)/\tau)},
\label{eq:pureGroup}
\end{equation}
where $\mathbf{G}$ is the set of group features. Since the number of groups is a certain small value, we can conduct group contrasting globally with all the group pairs.

\begin{wrapfigure}[22]{r}{0.6\linewidth}
	\vspace{-0.32in}
	\includegraphics[width=\linewidth]{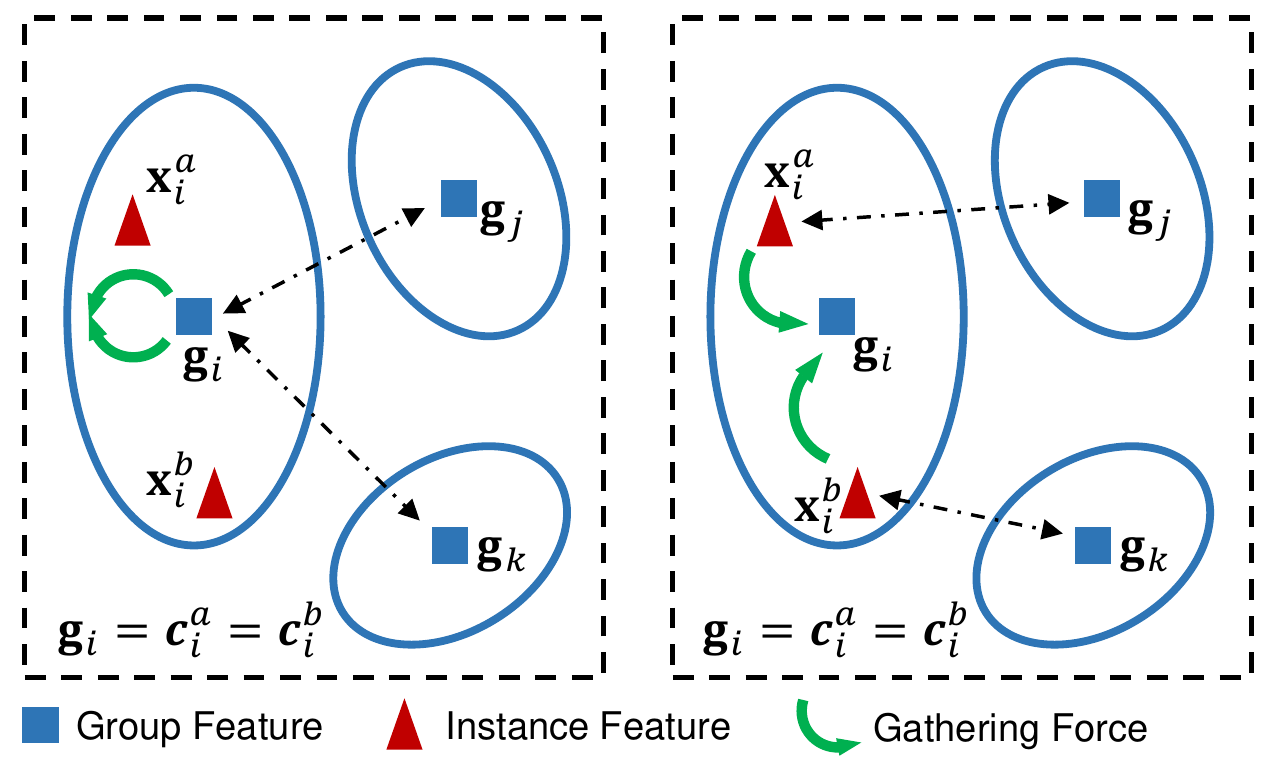}
	\caption{The left part illustrates the group contrastive target described by Eq.~\ref{eq:pureGroup}. As the training progresses, preliminary formed meaningful representations have a large chance to make $\mathbf{x}_i^a$ and $\mathbf{x}_i^b$ belong to the same group. In this case, the gathering force (the green arrow) cannot make them get closer. Thus, we modify it to the version in Eq.~\ref{eq:secondGroup} illustrated in the right part, where the push and pull force required by contrastive learning can always take effect.}
	\label{fig:exp_fail}
\end{wrapfigure}

The algorithm of gaining group features ($\phi$) will be detailed introduced in the next section. Here, we emphasize that besides well expressing a group of similar instances without extra limitation, a qualified $\phi$ also needs to make sure every group feature $\mathbf{c}_i$ is updated synchronously with the instance feature $f_\theta(\mathbf{x}_i)$ since the gradients need to be back-propagated through $\mathbf{c}_i$ to $f_\theta(\mathbf{x}_i)$ for training the parameters which means $\mathbf{c}_i$ can replace the latest $f_\theta(\mathbf{x}_i)$ to participate in contrasting. This is the core difference from the previous clustering-based method, since their group features gained by clustering cannot back-propagate the gradient, and directly contrasting group features cannot train the network.

Intuitively, the newly designed group contrastive learning method aims at directly adjusting the distribution of group features, and since gradients can propagate through group features to instance features, this algorithm can gradually learn the instance representation. However, unfortunately, it has flaws in the second half of the training process. With the meaningful representations gradually formed, $\mathbf{c}_i^a$ and $\mathbf{c}_i^b$ tend to be the same group feature. This will make the supervisory signal fail to gather similar instances and get them closer (see Fig.~\ref{fig:exp_fail}). To solve this problem, considering that $\mathbf{c}_i^a$ is the combination of a group of $f_\theta(\mathbf{x})$, we change the group contrastive learning loss to:
\begin{equation}
L_i = -{\rm log}\frac{{\rm exp}({\rm sim}(f_\theta(\mathbf{x}_i^a), \mathbf{c}_i^b)/\tau)}{\sum_{\mathbf{g}_j \in \mathbf{G}}{\rm exp}({\rm sim}(f_\theta(\mathbf{x}_i^a), \mathbf{g}_j)/\tau)}.
\label{eq:secondGroup}
\end{equation}
When $\mathbf{c}_i^a \neq \mathbf{c}_i^b$ Eq.~\ref{eq:secondGroup} can be seen as the ``stochastic-batch" version of Eq.~\ref{eq:pureGroup}, which splits an intact loss into several batches, similar to SGD and GD. When $\mathbf{c}_i^a = \mathbf{c}_i^b$, Eq.~\ref{eq:secondGroup} can still gather instances, which solves the problem. This formulation looks similar to previous clustering-based methods, but remember that we still conduct group contrasting instead of instance classification that previous methods do, since group features $\mathbf{c}_i$ and $ \mathbf{g}_j$ can propagate the gradient, thus has the same status with $f_\theta(\mathbf{x}_i^a)$ in contrasting and directly guide the learning direction.

\subsection{Synchronous Momentum Grouping}

\begin{figure}[t]
	\begin{center}
		\includegraphics[width=\linewidth]{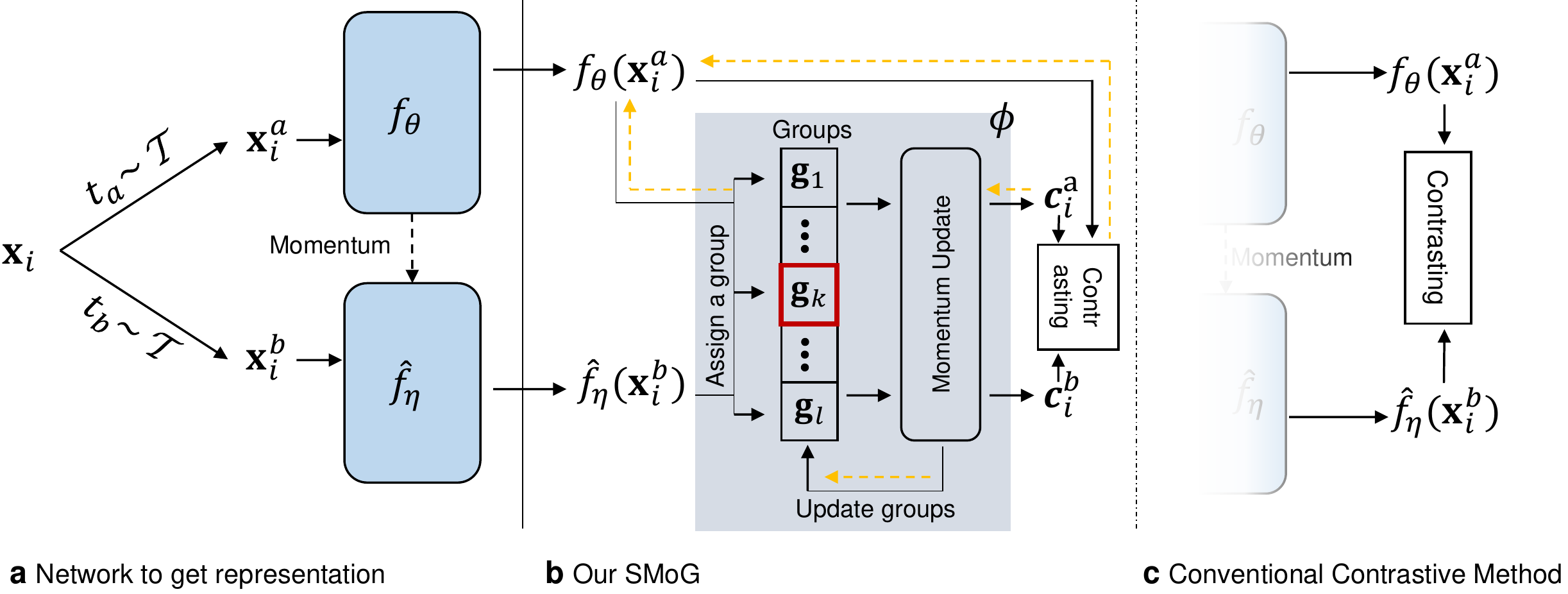}
	\end{center}
	\caption{Pipeline of our SMoG \textit{vs.} conventional contrastive method. Like typical contrastive learning methods, SMoG gets instance features from two augmentation views of the same image by a Siamese structure. Rather than directly contrasting the instance features, SMoG works at the group level. Specifically, it dynamically splits the already seen instances into groups, assigns the new instance to the closest group (the red one in the figure), and adopts the group feature which replaces the instance one to conduct the contrasting. Synchronous with the representation learning, group features are updated by the new instances in a momentum manner. The yellow dashed lines show the back-propagation paths, and we can see that the group feature can also back-propagate gradients.} 
	\label{fig:pipeline}
\end{figure}

SMoG's network structure and training settings are identical to conventional contrastive learning frameworks as Fig.~\ref{fig:pipeline} shows. That is generating $\mathbf{x}_i^a$ and $\mathbf{x}_i^b$ through two different groups of augmentation $t_a, t_b \sim \mathcal{T}$, and gaining their features through a Siamese network. Following MoCo~\cite{moco}, we adopt a momentum network: $\hat f_\eta \gets \alpha*\hat f_\eta + (1- \alpha) * f_\theta$, where $\alpha$ is the momentum ratio and here, $f_\theta$ denotes layers that also exist in $\hat f_\eta$. So far, SMoG is consistent with the typical contrastive learning methods. By adding the group assigning function $\phi$, we can get the complete SMoG method.

\subsubsection{Generating Group Features}
As mentioned in last section, group-level contrasting optimizes instance features through group features. Thus, group features must represent the latest instance ones to propagate gradients, namely, $\mathbf{c}_i$ needs to be synchronously updated with $f_\theta(\mathbf{x}_i)$ and calculated by a differentiable function. Directly adopting conventional global clustering as $\phi$ is not feasible to synchronously update $\mathbf{c}_i$ with $f_\theta(\mathbf{x}_i)$ at each iteration due to the computing cost. Thus, we simply modify it to the momentum grouping scheme, generating $\mathbf{c}_i$ through an iterative algorithm synchronized with representation learning.

Before the training start, we initialize the $l$ group features $\{\mathbf{g}_1, ..., \mathbf{g}_k, ..., \mathbf{g}_l\}$ randomly or using a clustering method such as k-means. Then along with the training process, we update $\mathbf{g}_k$ and get $\mathbf{c}_i$ in each iteration by:
\begin{equation}~\label{eq:mg}
\begin{aligned}
\mathbf{c}_i &= {\rm argmin}_{\mathbf{g}_k}({\rm sim}(f_\theta(\mathbf{x}_i), \mathbf{g}_k))\\
\mathbf{g}_k &\gets \beta * \mathbf{g}_k + (1 - \beta) * {\rm mean}_{\mathbf{c}_t=\mathbf{g}_k}{f_\theta(\mathbf{x}_t)},
\end{aligned}
\end{equation}
where $\beta$ is the momentum ratio, $\mathbf{x}_t$ comes from a mini-batch, and we omit the normalization. Importantly, this mechanism does not introduce extra limitation.

The momentum grouping scheme assigns each instance to the closest group and iteratively updates the group features in a momentum manner. In this way, the group features are always the latest representative of the instance visual features and more importantly, for each iteration, the updated $g_k$ is adopted to conduct contrasting and the gradient can be back-propagated through $g_k$ to $f_\theta(\mathbf{x}_i)$. This is the main difference from previous clustering-based methods. 
\subsubsection{Avoiding Collapse}
The momentum grouping scheme ensures that the group features $\mathbf{g}_k$ always represent the latest learned instance representations. However, because the algorithm is based on the iterative updating with a local subset of instances (a batch), at the early training stage, it might be unstable. The scale of the groups may be unbalanced or all the instances collapse into few groups. To deal with this, we periodically apply an extra grouping process on a cached relatively large feature set to relocate the groups. There will be long intervals among groupings which are light and quick, thus, the overhead can be ignored.

\subsection{Compared with Previous Clustering Methods}
Group discrimination is not a first proposed concept. Previous clustering based methods DeepClustering~\cite{deepcluster} and SwAV~\cite{swav} also conduct it and in terms of the loss function, the three algorithms have a similar form. The main difference lies on the method to generate the groups and the way to utilize them. Our SMoG aims at conducting the contrasting among the groups like Eq.~\ref{eq:pureGroup} shows. This is impossible for previous clustering-based methods since their group features have to be detached out and cannot back-propagate the gradients to the parameters. Thus, they do not contrast the groups but classify the instances into groups. To achieve the group contrasting, we propose the momentum grouping scheme which allows us directly contrasting the groups and propagate gradients. Although we modify the final loss to Eq.~\ref{eq:secondGroup} for better performance, our SMoG is still a group-level contrastive algorithm, since the group features directly guide the optimization direction. This is the reason we call our method "grouping contrasting" instead of "clustering" to distinguish them. SMoG conducts contrasting among groups instead of classifying instances based on clusters.

\subsection{Implementation Details}

\paragraph{Augmentation.}
We adopt the asymmetrical augmentation method used in BYOL \cite{byol}, where there are two augmentation schemes for the two streams of the Siamese network. The two schemes adopt the same color jittering but one has a stronger Gaussian blur and the other one has stronger solarization. Since the contrasting loss we adopt is also asymmetrical (contrasting among instance and group features), the two streams of the Siamese network are not assigned to a fixed augmentation scheme, instead, they adopt one of them in an alternate manner. In this way, the two streams can generate instance features under the same distribution so that they can be mapped to the same group feature set. 

\paragraph{SMoG Setting.}  
Like many previous works~\cite{byol,mocov3,moby}, both $f_\theta$ and $\hat f_\eta$ have a backbone and a projection head. And $\hat f_\eta$ has an extra prediction head stacked on the projection head. The backbone's output is adopted as the learning representation. The two kinds of head are both a two-layer MLP and their hidden layer is followed by a BatchNorm~\cite{bn} and an activation function (ReLU for ResNet and GELU~\cite{gelu} for SwinTransformer~\cite{SwinT}). The output layer of projection head also has BN. The dimension of the hidden layer is 2048, while for the output layer, it is 128 for CNN and 256 for Transformer. The Siamese's momentum ratio $\alpha$ is 0.999. In experiments, we group all the instances into 3k groups and the group features $\mathbf{g}$ are initialized with the K-means algorithm. The momentum ratio $\beta$ of grouping follows a linear schedule from 1.0 to 0.99. To avoid collapse, we reset the group features every 300 iterations with K-means on cached features of the past 300 iterations and $\hat f_\eta$ is synchronously reset with the parameters of $f_\theta$.

\paragraph{Pre-Training Details.}
We pre-train the models on ImageNet dataset~\cite{imagenet}. For CNN backbones, we train with the LARS~\cite{LARS} optimizer with 4096 mini-batchsize on 64 GPUS (when adopting ResNet50). The base learning rate is set to $lr = 0.3 \times {\rm batchsize} / 256$, following first a 10-epoch warm-up and then the cosine scheduler. The weight decay and temperature $\tau$ are set to $10^{-6}$ and 0.1. For Transformer based backbones, we adopt the Adamw optimizer~\cite{loshchilov2018fixing}. The base learning rate and weight decay are $5e^{-4} \times {\rm batchsize} /2048$ and 0.05. Other settings are the same with CNN backbones. For efficient training, we also adopt the multi-crop training scheme~\cite{swav,dino} with two large views (224$\times$224) and 4 small views (96$\times$96). When adopting multi-crop, the scale of the random crops are [0.2, 1.0] and [0.05, 0.2] respectively for large and small views.

\begin{wraptable}[29]{r}{0.55\linewidth}
	\vspace{-0.35in}
	\caption{Linear protocol results on ImageNet. ResNet50 is adopted. $\dagger$ denotes the model adopts the multi-crop training strategy. ``acc'' means accuracy.}
	\centering
	\setlength\arrayrulewidth{0.8pt}
	\resizebox{0.55\columnwidth}{!}{
		\begin{tabular}{l|cccc}
			model & epoch & batchsize & top1 acc & top5 acc\\
			\hline
			supervised & 100ep & 256 & 76.1 & 92.7 \\
			\hline
			\multicolumn{3}{l}{\textit{instance contrastive method}}\\
			ReSSL~\cite{ressl} & 200 & 256 & 69.6 & - \\
			MoCoV2~\cite{moco} & 800 & 256 & 71.1 & 90.1\\
			SimSiam~\cite{simsiam} & 800 & 256 & 71.3 & -\\
			InfoMin Aug.~\cite{infomin} & 800 & 256 & 73.0 & 91.1\\
			MoCoV3~\cite{mocov3} & 400 & 4096 & 73.1 & -\\
			MoCoV3 & 800 & 4096 & 73.8 & -\\
			BYOL~\cite{byol} & 400 & 4096 & 73.2 & -\\
			BYOL & 800 & 4096 & 74.3 & 91.6\\
			Barlow Twins~\cite{barlow} & 1000 & 2048 & 73.2 & 91.0\\
			RELIC~\cite{Relic} & 1000 & 4096 & 74.8 & 92.2\\
			SSL-HSIC~\cite{HSIC} & 1000 & 4096 & 74.8 & 92.2\\ 
			
			\hline
			\multicolumn{3}{l}{\textit{group-based method}}\\
			DeepCluster & 400 & 256 & 52.2 & -\\
			ODC~\cite{ODC} & 400 & 256 & 57.6 & -\\
			PCL~\cite{pcl} & 200 & 256 & 67.6 & -\\
			DeepClusterV2 & 400 & 4096 & 70.2 & -\\
			SwAV & 400 & 4096 & 70.1 & -\\
			CoKe~\cite{coke} & 800 & 1024 & 72.2 & -\\
			\rowcolor[gray]{0.95} SMoG & 400 & 2048 & 73.6 & 91.3\\
			\rowcolor[gray]{0.95} SMoG & 800 & 4096 & 74.5 & 91.9\\
			\hline
			\multicolumn{3}{l}{\textit{with multi-crop}}\\
			SwAV$\dagger$ & 400 & 4096 & 74.6 & -\\
			SwAV$\dagger$ & 800 & 4096 & 75.3 & -\\
			DC-v2$\dagger$ & 800 & 4096 & 75.2 & - \\
			DINO $\dagger$~\cite{dino} & 800 & 4096 & 75.3 & - \\
			UniGrad $\dagger$~\cite{unigrad} & 800 & 4096 & 75.5 & - \\
			NNCLR $\dagger$~\cite{nnclr} & 1000 & 4096 & 75.6 & 92.4\\ 
			\rowcolor[gray]{0.95} SMoG$\dagger$ & 400 & 4096 & \textbf{76.4} & \textbf{93.1}\\
	\end{tabular}}
	\label{tab:res50_res}
\end{wraptable}

\section{Experiments}
We evaluate SMoG firstly on the standard benchmark on ImageNet with both CNN and Transformer backbones. We then compare the performance on several downstream tasks and give a detailed ablation study.

\subsection{Linear Evaluation}
Following the standard benchmarks~\cite{moco,simclr,byol,swav,dino,mocov3,moby}, we first evaluate the representation of ResNet-50~\cite{resnet} and Swin Transformer Tiny~\cite{SwinT} trained with the proposed SMoG by the linear protocol: linear classification on frozen features.

Results on ResNet-50 is shown in Tab.~\ref{tab:res50_res} and we can see that previous group-based methods have no advantages over instance contrastive ones (about -2\% top-1 accuracy) due to the introduced restrictions. While the proposed SMoG improves the best performance of group-based methods by 2.3\% top-1 accuracy and achieves the SOTA performances among all the unsupervised representation methods. More importantly, after adopting the multi-crop training strategy~\cite{swav,dino}, for the first time, the unsupervised representation of ResNet-50 surpasses the performance of the vanilla supervised one (+ 0.3\%) on ImageNet. 

We adopt Swin-Transformer to evaluate the performance of SMoG on Transformer backbones. Results are shown in Tab.~\ref{tab:trans_res}. For the two-view setting, most Transformer-based backbones perform worse than the CNNs with similar parameters. But Transformers benefit more from the multi-view training. We hold
\begin{wraptable}[20]{r}{0.6\linewidth}
	\caption{Linear protocol results on ImageNet. $\dagger$ denotes adopting the multi-crop training strategy. The throughput (im/s) is calculated on a NVIDIA V100 GPU with 128 samples per forward. We report performances without Mixup and standard ones for supervised methods. }
	\centering
	\setlength\arrayrulewidth{0.8pt}
	\resizebox{0.6\columnwidth}{!}{
		\begin{tabular}{l|cccc}
			model & backbone & throughput &param & top1 acc\\
			\hline
			supervised & SwinT & 808 & 28 & 77.8 / 81.3\\
			Supervised & DeiT-S/16 & 1007 & 21 & 77.5 / 79.8\\
			\hline
			MoBY & DeiT-S/16 & 1007 & 21 & 72.8\\
			MoBY & SwinT & 808 & 28 & 75.0\\
			MoCoV3 & DeiT-S/16 & 1007 & 21 & 72.5\\
			MoCoV3 & ViT-B/16 & 312 & 85 & 76.7 \\
			DINO & DeiT-S/16 & 1007 & 21 & 72.5\\
			DINO $\dagger$ & DeiT-S/16 & 1007 & 21 & 77.0\\
			EsViT~\cite{li2021efficient} & SwinT & 808 & 28 & 70.5\\ 
			EsViT $\dagger$& SwinT & 808 & 28 & 77.0\\ 
			\rowcolor[gray]{0.95} SMoG & SwinT & 808 & 28 & 74.5\\
			\rowcolor[gray]{0.95} SMoG $\dagger$ & SwinT & 808 & 28 &  \textbf{77.7}\\
	\end{tabular}}
	\label{tab:trans_res}
\end{wraptable}
the opinion that the self-attention, a global operator, needs more data to train. Thus, stronger augmentation leads to better performances. SMoG, in the two-view setting, achieves SOTA performances compared to DINO (+2\%), MoCo (+2\%), EsViT (+4\%), and MOBY (-0.5\%). Note that, for fair comparison, we report the performance of EsViT with only $\mathcal{L}_V$. In the multi-crops setting, SMoG achieves the supervised performance without Mixup~\cite{mixup} augmentation. Since mixup needs labels, it is not straightforwardly suitable for the unsupervised setting. This comparison reveals a landmark progress in contrastive learning.

\begin{wraptable}[14]{r}{0.6\linewidth}
	\vspace{-0.35in}
	\caption{Linear protocol results on ImageNet with larger backbones. We experiment on wider ResNet. The pre-training details are the same with the ResNet-50 $\times$1.}
	\centering
	\setlength\arrayrulewidth{0.8pt}
	\resizebox{0.6\columnwidth}{!}{
		\begin{tabular}{l|cccc}
			model  & backbone & param & top1 acc & top5 acc\\
			\hline		 
			\rowcolor[gray]{0.95}  & Res50 (x2) & 188 & 77.8 & 93.8\\
			\rowcolor[gray]{0.95} \multirow{-2}{*}{Supervised} & Res50 (x4) & 375& 78.9 & 94.5 \\
			\multirow{2}{*}{BYOL} & Res50 (x2) & 188 & 77.4 & 93.6\\
			& Res50 (x4) & 375& 78.6 & 94.2 \\
			\rowcolor[gray]{0.95} & Res50 (x2) & 188 & 77.3 & -\\
			\rowcolor[gray]{0.95} \multirow{-2}{*}{SwAV}  & Res50 (x4) & 375& 77.9 & - \\
			\multirow{2}{*}{SMoG} & Res50 (x2) & 188 & 78.0 & 93.9\\
			& Res50 (x4) & 375 & 79.0 &  94.4\\
			
	\end{tabular}}
	\label{tab:largeArch}
\end{wraptable}

\begin{SCfigure}
	\includegraphics[width=0.55\linewidth]{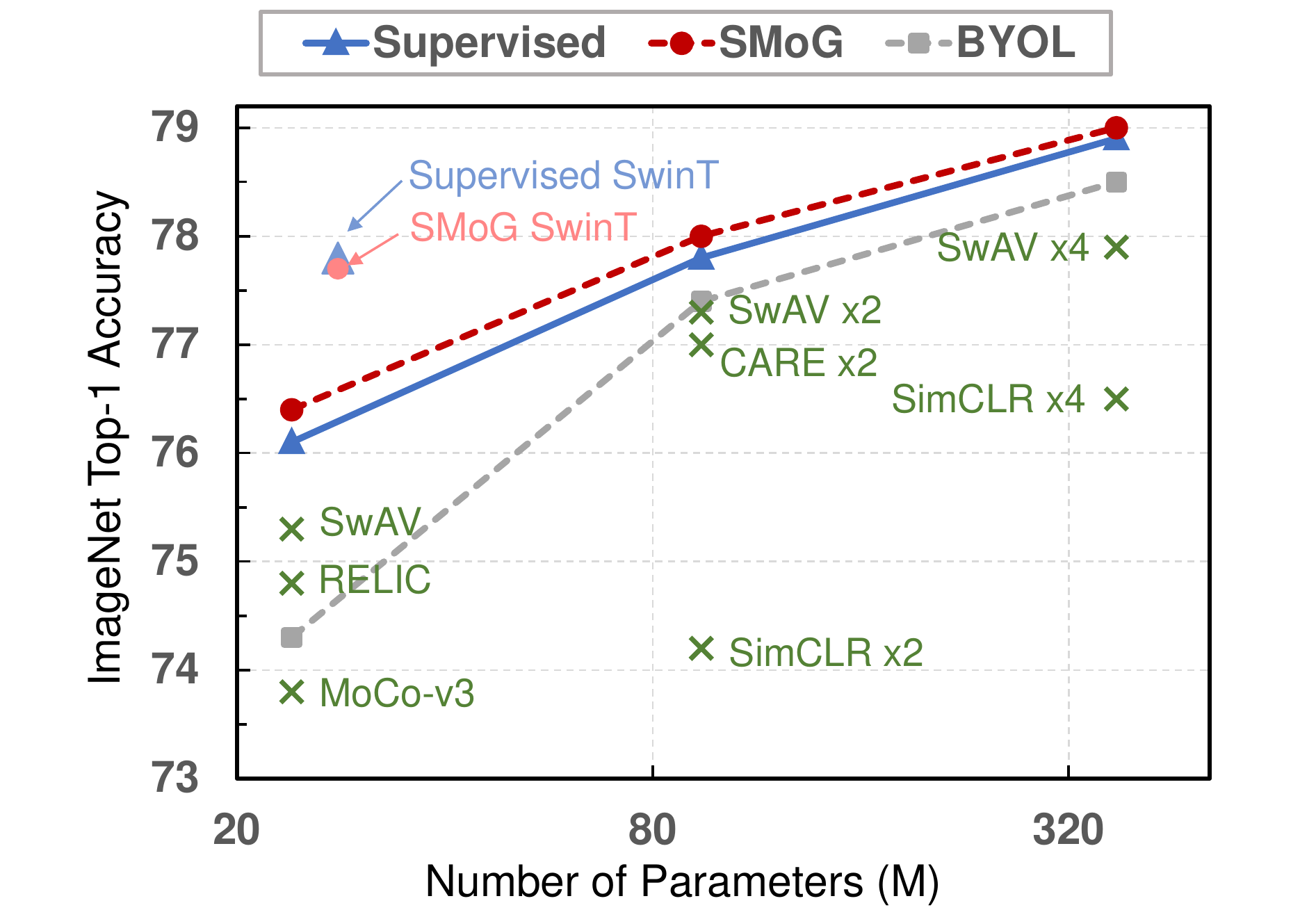}
	\caption{ResNet linear evaluation on ImageNet compared with supervised training. We can see that for different model sizes, our SMoG achieves comparable performance with supervised method.}
	\label{fig:compare_w_sup}
\end{SCfigure}

\paragraph{Large Architectures}
Tab.~\ref{tab:largeArch} shows the results on several varients of ResNet-50 with larger widths~\cite{kolesnikov2019revisiting}. The performance increases with a similar trend of supervised method and previous unsupervised framework. And it is worth noting that in previous works~\cite{simclr,byol}, with larger architectures, their gap with the supervised learning decreases, but it is a pity that we do not observe an increasing superiority of SMoG over the supervised one. In Fig.~\ref{fig:compare_w_sup}, we demonstrate that on both CNN and Transformer, SMoG is comparable with the vanilla supervised method.

\subsection{Semi-Supervised Fine-Tune Evaluation}

Next, we evaluate the proposed SMoG through semi-supervised fine-tuning the unsupervised representation on a subset of training data of ImageNet. We follow the protocol adopted in \cite{simclr} and the 1\% and 10\% labeled splits of ImageNet we adopt are the fixed ones provided in \cite{simclr}.

\begin{wraptable}[16]{r}{0.6\linewidth}
	\vspace{-0.05in}
	\caption{Semi-supervised results on ImageNet.}
	\centering
	\setlength\arrayrulewidth{0.8pt}
	\resizebox{0.58\columnwidth}{!}{
		\begin{tabular}{l|ccc|ccc}
			\multirow{2}{*}{method} & \multicolumn{3}{c|}{Top-1 Acc (\%)}&\multicolumn{3}{c}{Top-5 Acc (\%)}\\
			& 1\% & 10\% & 100\% & 1\% & 10\% & 100\% \\
			\hline
			\multicolumn{7}{l}{\textit{resnet-50 $\times$1}}\\
			Supervised & 25.4 & 56.4 & 76.1 & 48.4 & 80.4 & 92.9\\
			SimCLR & 48.3 & 65.6 & 76.0 & 75.5 & 87.8 & 93.1\\
			BYOL & 53.2 & 68.8 & 77.7 & 78.4 & 89.0 & 93.9\\
			SwAV & 53.9 & 70.2 & - & 78.5 & 89.9 & -\\
			Barlow Twins & 55.0 & 69.7 & - & 79.2 & 89.3 & -\\
			SSL-HSIC & 52.1 & 67.9 & 77.2 & 77.7 & 88.6 & 93.6\\
			NNCLR & 56.4 & 69.8 & - & 80.7 & 89.3 & - \\
			\rowcolor[gray]{0.95} SMoG & \textbf{58.0} & \textbf{71.2} & \textbf{78.3} & \textbf{81.6} & \textbf{90.5} & \textbf{94.2}\\
			\hline
			\multicolumn{7}{l}{\textit{resnet-50 $\times$2}}\\
			Supervised & - & - & 77.8 & - & -& 93.8\\
			SimCLR & 58.5 & 71.7 & - & 83.0 & 91.2 &  -\\
			BYOL & 62.2 & 73.5 & - & 84.1 & 91.7 &-\\
			\rowcolor[gray]{0.95} SMoG & \textbf{63.6} & \textbf{74.4} & \textbf{80.2} & \textbf{85.6} & \textbf{92.4} & \textbf{95.2}\\
	\end{tabular}}
	\label{tab:semisup_res}
\end{wraptable}

As shown in Tab.~\ref{tab:semisup_res}, SMoG consistently outperforms all the previous methods in both 1\% and 10\% settings on ResNet-50 $\times$1 and $\times$2. Besides, we also fine-tune the unsupervised representation on the full ImageNet. With SMoG pre-training, ResNet-50 achieves 78.3\% top-1 accuracy under the standard training recipe, surpassing the directly supervised learning by 2.2\%. Similarly, the ResNet-50 $\times$ 2 also outperforms the directly supervised training by 2.4\%.

\subsection{Transfer to Other Vision Tasks}
\begin{wraptable}[12]{r}{0.6\linewidth}
	\vspace{-0.35in}
	\caption{Transfer learning results on semantic segmentation task. We fine-tune the representations on VOC2012 and Cityscapes dataset. The segmentation model is FCN with ResNet-50.}
	\small
	\centering
	\setlength\arrayrulewidth{0.8pt}
	\begin{tabular}{l|cc|cc}
		\multirow{2}{*}{model} & \multicolumn{2}{c|}{Cityscapes}&\multicolumn{2}{c}{VOC-2012}\\
		& mIoU & mAcc & mIoU & mAcc \\
		\hline
		Supervised & 73.83 & 82.56 & 73.59 & 83.74 \\
		MoCoV2 & 74.30 & 83.37 & 70.86 & 80.37 \\
		SwAV & 74.80 & 83.01 & 74.97 & 84.27\\
		BYOL & 74.90 & 83.73 & 74.76 & 84.37\\
		\rowcolor[gray]{0.95} SMoG & \textbf{76.03} & \textbf{83.97} & \textbf{76.22} & \textbf{85.01}\\
		
	\end{tabular}
	\label{tab:seg_res}
\end{wraptable}
We further transfer the unsupervised representations of ResNet-50 learned with the proposed SMoG to several downstream tasks, namely semantic segmentation, object detection, and instance segmentation.

We first evaluate SMoG on Cityscapes~\cite{cityscapes} and VOC-2012~\cite{voc} semantic segmentation tasks. For fair comparison, we align all the methods with the same training recipe of FCN~\cite{fcn}. Results of mean accuracy and mean IoU are provided in Tab.~\ref{tab:seg_res}, we can see that nearly all the unsupervised representations outperform the conventional supervised one which reveals that in this downstream task, unsupervised representation is already a better choice. And again, SMoG performs better than the original SOTAs on Cityscapes (+1.1 mIoU) and VOC-2012 (+1.4 mIoU) datasets.

Then, we evaluate on object detection and instance segmentation tasks on COCO~\cite{coco} dataset. Similarly, all the unsupervised representations are transferred with the same fine-tuning recipe of Mask-RCNN~\cite{maskrcnn}. We provide AP results in Tab.~\ref{tab:det_res}. The same with semantic segmentation, on these two downstream tasks, our unsupervised pre-training representations are better than the supervised one. And still, SMoG improves the current SOTA results, +0.7 AP on object detection and +1.1 AP on instance segmentation.

\subsection{Ablation Study}
\begin{wraptable}[15]{r}{0.6\linewidth}
	\vspace{-0.35in}
	\caption{Transfer learning results on object detection and instance segmentation tasks. we adopt COCO as the fine-tuning dataset. Mask RCNN with ResNet-50-FPN is the detection and segmentation model. We report the AP metrics.}
	\centering
	\setlength\tabcolsep{4pt}
	\resizebox{0.6\columnwidth}{!}{
		\begin{tabular}{l|ccc|ccc}
			& \multicolumn{3}{c|}{COCO det} & \multicolumn{3}{c}{COCO instance seg.}\\
			method & AP$^{bb}$ & AP$^{bb}_{50}$ & AP$^{bb}_{75}$ & AP$^{mk}$ &AP $^{mk}_{50}$ & AP$^{mk}_{75}$\\
			\hline
			Rand Init & 31.0 & 49.5 & 33.2 & 28.5 & 46.8 & 30.4 \\
			Supervised & 38.9 & 59.6 & 42.7 & 35.4 & 56.5 & 38.1 \\
			\hline
			InsDis~\cite{wu2018unsupervised} & 37.4 & 57.6 & 40.6 & 34.1 & 54.6 & 36.4 \\
			PIRL~\cite{misra2020self} & 38.5  & 57.6 & 41.2 & 34.0 & 54.6 & 36.2 \\
			MoCoV2 & 39.4 & 59.9 & 43.0 & 35.8 & 56.9 & 38.4 \\
			SwAV & 38.5 & 60.4 & 41.4 & 35.4 & 57.0 & 37.7 \\
			DC-v2~\cite{swav} & 38.3 & 60.3 & 41.3 & 35.4 & 56.7 & 38.0 \\
			BYOL & 39.4 & 59.9 & 43.0 & 35.8 & 56.8 & 38.5\\
			Barlow Twins & 39.2 & 58.7 & 42.6 & 34.3 & 55.4 & 36.5 \\
			\rowcolor[gray]{0.95} SMoG & \textbf{40.1} & \textbf{61.6} & \textbf{43.7} & \textbf{36.9} & \textbf{58.7}  & \textbf{39.3} \\
			
	\end{tabular}}
	\label{tab:det_res}
\end{wraptable}
We provide ablation studies on key components of SMoG. ResNet-50 is adopted. Network is trained for 100 epochs without multi-crop. The representation is evaluated under the linear protocol.

\begin{figure}
	\begin{center}
		\includegraphics[width=0.6\linewidth]{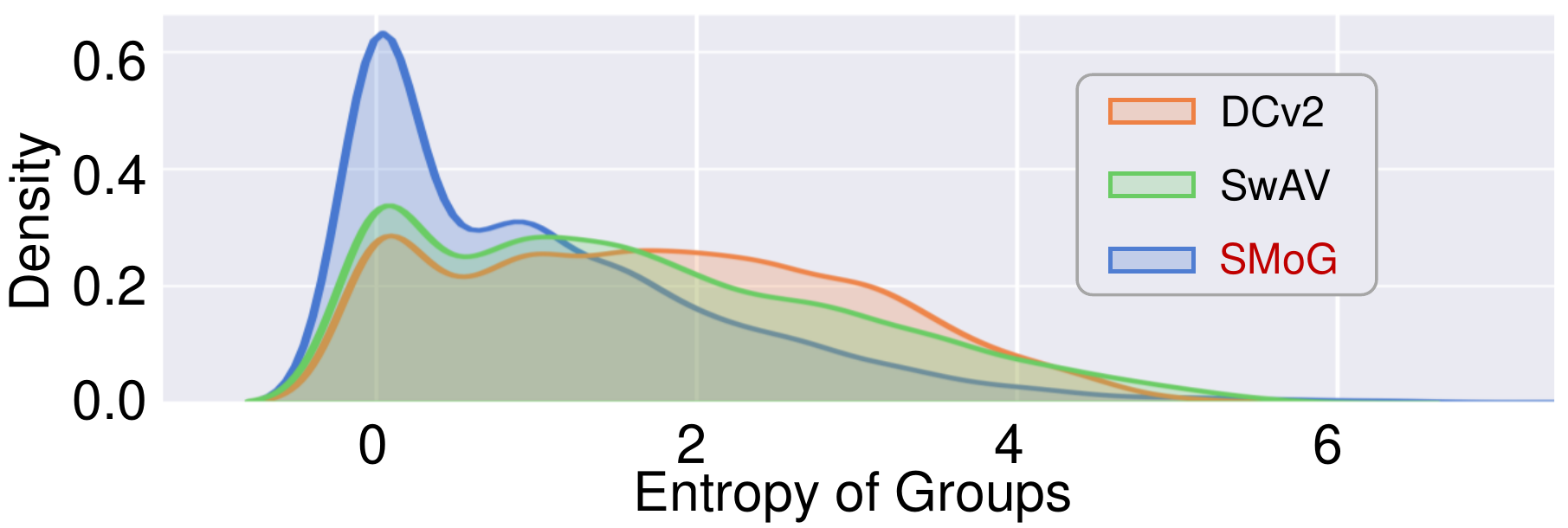}
	\end{center}
	\caption{Entropy of each groups. SMoG produces groups with much lower entropy which represents a better grouping quality in terms of high-level semantics.}
	\label{fig:entropy}
\end{figure}

\paragraph{Grouping quality} To evaluate the grouping quality, in Fig.~\ref{fig:entropy}, we provide the density distribution of group entropy. All the three group-discrimination methods (our SMoG, SwAV, and DeepClusteringv2) have 3k groups. For each group, its entropy is $\sum_i -p_i*{\rm log}(p_i)$ where $p_i$ is the ratio of instances that belong to class $i$ (data annotation) in this group. A lower entropy means a group has more unitary semantics and is more meaningful. Thus, more groups with low entropy indicate higher grouping quality. From Fig.~\ref{fig:entropy}, we can see SMoG has twice more low-entropy groups than the two baselines, proving its superiority.

\paragraph{Momentum ratio $\beta$ of group features}
From Tab.~\ref{tab:abl-beta}, we can see that the linearly decreasing schedule for $\beta$ is much better than the fixed schedule. This is because our momentum grouping scheme updates the group features with only a small part of features (a batch) in each iteration and at the beginning of training, the feature distribution changes drastically, a small $\beta$ will lead to unstable groups. Thus, we need a large $\beta$ at the beginning to solve the problem and the linear schedule is a straightforward solution. After adopting the linearly decreasing schedule, our SMoG is not sensitive to the final value of $\beta$ and we adopt 0.99 as the default value for all the experiments on both CNN and Transformer.

\paragraph{Avoiding collapse}
Still, because the group feature updating process cannot access the global feature distribution, there is a possibility of collapse during training. Thus, we adopt the periodical clustering (pd) trick to avoid this. The first two rows of Tab.~\ref{tab:abl-trick} prove its necessity. We can see that it successfully makes the backbone learn useful representations but the performance is not good enough. We believe this is because the group features are always generated and updated based on the instance features generated by only $f_\theta$, leading to misalignment with features from $\hat f_\eta$. Thus, we also reset $\hat f_\eta$ with the parameters of $f_\theta$ after each clustering. The integrated pd with $\hat f_\eta$ reset leads to ideal performance. And surprisingly, we find that only resetting $\hat f_\eta$ without the clustering can avoid the collapse too. We think the sudden change of $\hat f_\eta$ feature distribution also helps to avoid the network gradually falling into degeneration. 

\begin{table}
	\caption{Ablation study on SMoG. We adopt ResNet50 as the backbone and train the unsupervised algorithm on ImageNet for 100 epochs without multi-crop training strategy. We report the linear evaluation results (Top-1 accuracy).}
	\centering
	\setlength\tabcolsep{9pt}
	\begin{subtable}[t]{0.42\linewidth}
		\centering
		\caption{Momentum ratio $\beta$. Linearly decreasing schedule performs better.}
		\resizebox{\columnwidth}{!}{
			\begin{tabular}{l|c}~\label{tab:abl-beta}
				Schedule & Top1 acc\\
				\hline
				fixed 0.99 $\beta$ & 65.9\\
				1.0 $\beta$ $\rightarrow$ 0.9 $\beta$ & 67.0\\
				1.0 $\beta$$\rightarrow$0.99 $\beta$& 67.2\\
				1.0 $\beta$ $\rightarrow$0.999 $\beta$ & 67.1\\
				\hline
		\end{tabular}}
	\end{subtable}
	~
	\begin{subtable}[t]{0.55\linewidth}
		\centering
		\caption{Tricks dealing with collapse. The periodical clustering is necessary.}
		\label{tab:abl-trick}
		\resizebox{\columnwidth}{!}{
			\begin{tabular}{ l| c}
				Tricks& Top1 acc \\
				\hline
				None & 0.1\\
				+ periodical clustering (pd) & 53.7\\
				+ reset $\hat f_\eta$ periodically & 53.4\\
				+ pd \& reset $\hat f_\eta$ periodically & 67.2\\
				\hline
				
		\end{tabular}}
	\end{subtable}
	~~
	\begin{subtable}[t]{0.33\linewidth}
		\centering
		\caption{Number of groups. 3k is enough for SMoG.}
		\label{tab:abl-groups}
		\setlength\tabcolsep{7.8pt}
		\resizebox{\columnwidth}{!}{
			\begin{tabular}{ l|c }
				\# Groups & Top1 acc \\
				\hline
				300 & 65.2\\
				1000 & 67.0\\
				3000 & 67.2\\
				10000 & 67.2\\
				\hline
				
		\end{tabular}}
	\end{subtable}
	~
	\begin{subtable}[t]{0.45\linewidth}
		\centering
		\caption{Grouping algo. Momentum update better balances grouping and learning.}
		\label{tab:abl-method}
		\resizebox{\columnwidth}{!}{
			\begin{tabular}{ l|c }
				method & Top1 acc \\
				\hline
				Randomly select & 30.2\\
				Adopt latest & 42.1\\
				Averaging update & 65.8\\
				Momentum update & 67.2\\
				
				\hline
				
		\end{tabular}}
	\end{subtable}
	\label{tab:abl}
\end{table}

\paragraph{Number of groups}
In Tab.~\ref{tab:abl-groups}, we evaluate the influence of the number of groups used in SMoG under the linear protocol. The results show that SMoG is not sensitive to the group number. Even if we tune it in a wide range (1k$\sim$30k), the performance is maintained at a stable level ($\pm$ 0.2), as long as there are enough groups (300 groups are too few to perform well). This is consistent with the conclusion of SwAV~\cite{swav}. More groups increase the computation time consuming of the momentum grouping algorithm ($\phi$) since it needs node communication to synchronize the group features among all the nodes. Thus, we adopt 3K as the default setting for all the experiments.

\paragraph{Method for updating group features}
In Tab.~\ref{tab:abl-method}, we evaluate different operators for updating the group features. We consider 4 operators here: 1) Randomly select (RS): randomly select a group of latest instance features and adopt them as the group features. 2) Adopt latest (AL): $\mathbf{g}_k \gets {\rm mean}_{\mathbf{c}_t=\mathbf{g}_k}{f_\theta(\mathbf{x}_t)}.$ 3) Averaging update (AU): 
$\mathbf{g}_k \gets \mathbf{g}_k + (1/n) * ({\rm mean}_{\mathbf{c}_t=\mathbf{g}_k}{f_\theta(\mathbf{x}_t)} - \mathbf{g}_k)$, where $n$ is the total number of instances belonging to group $k$, including ones in current iteration.
4) Momentum update (MU): the proposed method described in Eq.~\ref{eq:mg}.

As the random baseline, RS's poor performance shows us that the grouping algorithm plays an important role in our SMoG method. AL also does not perform well, which implies that the local grouping method will not provide an efficient training signal for representation learning. AU follows the sequential k-means algorithm and achieves a relatively great performance. But compared to our MU, it is not the best choice for conducting grouping together with the representation learning. The specifically designed MU which gives new features more weights is more suitable for the group-level contrastive learning.

\section{Conclusion}
We extend the recently popular contrastive learning to group level with the proposed SMoG and push the performance of unsupervised representation under linear protocol to the vanilla supervised level. SMoG synchronously conducts the representation learning and grouping process to effectively achieve the group-level contrasting. We hope the newly designed group-level contrastive learning will be useful for the community to further develop visual unsupervised method.

\paragraph{Acknowledgement}~ This work was supported by the National Key R\&D Program of China (2021ZD0110700), Shanghai Municipal Science and Technology Major Project (2021SHZDZX0102), Shanghai Qi Zhi Institute, and SHEITC (018-RGZN-02046).

\clearpage
%
%
\bibliographystyle{style/splncs04}
\bibliography{egbib}
\end{document}